\documentclass[sigconf]{acmart}

\AtBeginDocument{%
  \providecommand\BibTeX{{%
    \normalfont B\kern-0.5em{\scshape i\kern-0.25em b}\kern-0.8em\TeX}}}

\setcopyright{acmcopyright}
\copyrightyear{2022}
\acmYear{2022}
\acmDOI{XXXXXXX.XXXXXXX}

\acmConference[XXXXXXX]{}{XXXXXXX}{XXXXXXX}
\usepackage{multirow}
\usepackage[ruled,linesnumbered]{algorithm2e}
\usepackage{comment}
\usepackage{adjustbox}

\begin{document}

\title{Robust Task-Oriented Dialogue Generation with Contrastive Pre-training and Adversarial Filtering}

\author{Shiquan Yang, Xinting Huang, Jey Han Lau, Sarah Erfani \\
The University of Melbourne, Australia\\
\{shiquan@student., xintingh@student., laujh@, sarah.erfani@\}unimelb.edu.au
}

\renewcommand{\shortauthors}{Shiquan Yang, Xinting Huang, Jey Han Lau, Sarah Erfani}



\begin{abstract}
Data artifacts incentivize machine learning models to learn non-transferable generalizations by taking advantage of shortcuts in the data, and
there is growing evidence that data artifacts play a role for the strong results that 
deep learning models  achieve in recent natural language processing benchmarks.
In this paper, we focus on task-oriented dialogue and investigate  whether popular 
datasets such as MultiWOZ contain such data artifacts.
We found that by only keeping frequent phrases in the training
examples, state-of-the-art models perform similarly compared 
to the variant trained with full data, suggesting 
they exploit these spurious correlations
to solve the task. Motivated by this, we propose 
a contrastive learning based framework to encourage the model 
to ignore these cues and focus on learning generalisable patterns. We also experiment with adversarial filtering to remove ``easy'' training instances so that the model would focus on learning from the ``harder'' instances. We conduct a number of generalization
 experiments --- e.g., cross-domain/dataset and adversarial tests --- to assess the robustness of our approach and found that it works exceptionally well.

\end{abstract}

\maketitle

\section{Introduction}
Task-oriented dialogue systems aim to help human accomplish 
certain tasks such as restaurant reservation or navigation via natural language utterances. 
Recently, pre-trained language models 
\cite{hosseini2020simple,peng2020soloist,wu2020tod} achieve 
impressive results on dialogue response generation and 
knowledge base (KB) reasoning, two core components of 
dialogue systems. However, neural networks 
are found to be prone to learning \textit{data artifacts} \cite{mccoy2019right,ilyas2019adversarial}, i.e.\ superficial statistical patterns  
in the training data, and as such these results may not generalise to more challenging test cases,
e.g., test data that is drawn from a different distribution to the training data.

This issue has been documented in several natural language 
processing (NLP) tasks \cite{branco-etal-2021-shortcutted,mccoy2019right,niven2019probing}.
For example, in natural language inference (NLI), where the 
task is to determine whether one given sentence entails the other, 
the models trained on NLI benchmark datasets are highly
likely to assign a 
``contradiction'' label if there exists a word \textit{not} in the 
input sentences even if the true relation is ``entailment'', 
as \textit{not} often co-occurs with the label 
``contradiction'' in the training set. Similar issues have also 
been observed in many other tasks such as commonsense reasoning \cite{branco-etal-2021-shortcutted}, 
visual question answering \cite{qi2020causal,niu2021counterfactual}, and argument reasoning
\cite{niven2019probing}. However,
it's unclear whether such shortcuts exist 
in popular task-oriented dialogue datasets such as MultiWOZ \cite{eric2019multiwoz},
and whether existing dialogue models are genuinely learning the
underlying task or exploiting biases\footnote{We use the terms \textit{artifacts}, \textit{biases}, \textit{cues} and \textit{shortcuts} to denote the same concept and  use them interchangeably throughout the paper.} hidden in the data.

To investigate this, we start by probing whether 
state-of-the-art dialogue models are discovering and
exploiting spurious correlations on a popular task-oriented 
dialogue dataset. 
Specifically, we measure two
 state-of-the-art dialogue models' performance under two different configurations: 
full input (original dialogue history, e.g., \textit{I need to find a moderately priced hotel})
and partial input (dialogue history that contains only frequent phrases, e.g., \textit{I need to}). Preliminary experiments found that these models perform similarly under the two configurations, 
suggesting that these models have picked up these 
cues --- frequent word patterns which are often not meaning bearing --- to make predictions. This implies that 
these models did not learn transferable generalizations 
for the task, and will likely perform poorly on \textit{out-of-distribution} test data, e.g.,
one that has a different distribution to the training data.

To address this, we decompose task-oriented dialogue into two task: delexicalized response generation and KB reasoning (prediction of the right entities in the response), and explore methods to improve model robustness for the latter.
Using frequent phrases as the basis of dataset bias, 
 we experiment with contrastive learning to encourage the model to ignore these phrases to focus on meaning bearing words. Specifically, we pre-train our language model with a contrastive objective to encourage it to learn a similar representation for an original input (e.g., \textit{I need to find a moderately priced hotel}) and its debiased pair (e.g., \textit{find a moderately priced hotel})
before fine-tuning it for KB entity prediction.

Another source of bias comes from the data distribution  \cite{branco-etal-2021-shortcutted}. We found that the KB entity distribution in MultiWOZ can be highly skewed in certain contexts, e.g., if the dialogue context starts with \textit{I need to}, the probability of the KB entity \textit{Cambridge}  substantially exceeds chance level, which leads to inadequate learning of entities in the tail of the distribution. Here we adapt an adversarial filtering algorithm 
\cite{sakaguchi2019winogrande} to our task, which filters ``easy samples'' (i.e., samples in the head of the distribution) in the 
training data to create a more balanced data distribution so as to encourage the model to learn from the tail of the distribution.

We conduct a systematic evaluation on the robustness of our method and four state-of-the-art  task-oriented dialogue systems under various 
out-of-distribution settings. Experimental results demonstrate
that our method substantially outperforms these benchmark systems.

To summarize, our contributions are as follows:

\begin{itemize}
    \item We conduct analysis on a popular task-oriented 
    dialogue dataset and reveal shortcuts based on
    frequent word heuristics.
    \item We propose a two-stage contrastive learning framework 
    to debias spurious cues in the model inputs, and adapt adversarial filtering to create a more balanced training data distribution to improve the robustness of our task-oriented dialogue system.
    \item We perform comprehensive experiments to validate the robustness of our method against a number of strong benchmark systems in various 
    out-of-distribution test settings and found our method substantially outperforms its competitors.
\end{itemize}

\section{Related Work}
\subsection{Task-oriented Dialogue}
Traditionally, task-oriented dialogue systems are 
built via pipeline based approach where four 
independently designed and trained modules are connected 
together to generate the final system responses. These include
natural language understanding \cite{chen2016end} , dialogue state 
tracking \cite{wu2019transferable,zhong2018global}, 
policy learning \cite{peng2018deep}, and natural language 
generation \cite{chen2019semantically}. However, the 
pipeline based approach can be very costly and time-consuming as 
each module needs module-specific training data and can not be 
optimized in a unified way. To address this, many end-to-end 
approaches \cite{bordes2016endtoendtaskorientedlearning,lei-etal-2018-sequicity,mem2seq} have been proposed to reduce human efforts
in recent years. \citet{lei-etal-2018-sequicity} propose a two-stage 
sequence-to-sequence model to incorporate dialogue state tracking and 
response generation jointly in a single sequence-to-sequence architecture.
\citet{wu2019global} propose a global-local pointer mechanism that 
trains a global pointer at encoding stage using multi-task learning 
with addition supervisions extracted from the system responses. 
\citet{qin-etal-2020-dynamic} use a shared-private sequence-to-sequence 
architecture to handle the transferability of the system on 
KB incorporation and response generation. 
\citet{zhang2020task} 
propose a domain-aware multi-decoder network to combine belief state 
tracking, action prediction and response generation in a 
single neural architecture. More recently, the field has shifted 
towards using large-scale pre-trained language models such as 
BERT \cite{devlin2019bert} and GPT-2 \cite{radford2019language} for task-oriented 
dialogue modeling due to their success on many NLP tasks \cite{wolf2019transfertransfo,zhang2019dialogpt,peng2020soloist}.
\citet{peng2020soloist} and \citet{hosseini2020simple} employed a 
GPT-2 based model jointly trained for belief state prediction and 
response generation in a multi-task fashion. \citet{wu2020tod} pre-train 
BERT on multiple task-oriented dialogue datasets for response selection.

\subsection{Spurious Cues in NLP data}
Deep neural networks have achieved tremendous progress 
on many NLP tasks \cite{huang2019knowledgegraphembeddingbasedquestionanswering,ghazvininejad2018knowledgegrounded,xing2018hierarchicalrecurrentattentionnetwork,velivckovic2017graphattentionnetwork,wu2016ask} with the emergence of large-scale pre-trained language models 
like BERT \cite{devlin2018bert} and GPT-2 \cite{radford2019language}.
However, many recent NLP studies \cite{branco-etal-2021-shortcutted,mccoy2019right,niven2019probing,ilyas2019adversarial,hendrycks2021natural} have found 
that deep neural networks are prone to exploit spurious artifacts 
present in the data rather than learning the underlying task. In 
natural language inference (NLI), \cite{belinkov2019dont} found 
that certain linguistic phenomenon in the NLI benchmark
datasets correlate well with certain classes. For example,
by only looking at the hypothesis, simple classifier models
can perform as well as the model using full inputs (both hypothesis
and premise).
\cite{niven2019probing} found that BERT achieves a performance close to human on Argument Reasoning Comprehension Task (ARCT) 
with 77\% accuracy (3\% below human performance). However, they discover that 
the impressive performance is attributed to the exploitation 
of shortcuts in the dataset. \cite{geva2019modeling} analyze 
 annotator bias on NLP datasets 
and found that 
a model that uses only annotator identifiers can achieve a similar performance to one that uses the full data. In commonsense reasoning, \cite{branco-etal-2021-shortcutted} has 
performed a systematic investigation over four commonsense
related tasks and found that most datasets experimented with 
are problematic with models are prone to leveraging the 
non-robust features in the inputs to make decisions and 
do not generalize well to the overall tasks intended to 
be conveyed by the commonsense reasoning tasks and datasets.
Inspired by these studies, our paper focus on analyzing data artifacts in task-oriented dialogue datasets.

\subsection{Bias Reduction in NLP}
To address the data artifact issue, several approaches have 
been proposed.
In NLI, \citet{he2019unlearn} propose to train a debiased classifier 
by fitting the residuals of a biased classifier trained 
using insufficient features. However, their approach relies on task-specific
prior knowledge about the bias types such as hypothesis-only bias 
in NLI. \citet{sanh2020learning} propose to leverage a weak learner to 
automatically identify the biased examples in the training data 
and only use the hard examples to train the main model to obtain 
a debiased classifier. In commonsense reasoning, \citet{sakaguchi2019winogrande}
propose AF-Lite that iteratively removes 
``easy'' samples in the training data 
during model training to make model focus on the ``hard'' examples 
to improve the robustness of commonsense reasoning models. Note though, that
their approach is designed for classification tasks, and as such is not straightforward 
to adapt them to task-oriented dialogue. We tackle this by decomposing the problem into the two sub-tasks: lexicalized dialogue generation and KB entity prediction, so that we can apply AF-lite to the latter (as it is a classification problem).

\section{Spurious Cues In Task-oriented Dialogue Dataset}
\label{sec:preliminary}
To unveil potential linguistic artifacts in task-oriented dialogue 
datasets, we first conduct an investigation on MultiWOZ \cite{budzianowski2018multiwoz}, 
which is widely used among task-oriented dialogue
studies. By comparing the performance of a model trained using full dialogue history (e.g., \textit{I need to find a moderately priced hotel})
and partial history containing only frequent phrases (e.g., \textit{I need to}), it tells us if shortcuts exist in the dataset and the model has picked them up to solve the task. Note that these frequent phrases tend to be function words that don't bear much meaning (e.g., \textit{I need to}), and as such a model that can perform the task well using only them means it has not truly solved the task by capturing the underlying semantics of user utterances.

\begin{table}[]
    \centering
    \scalebox{1.0}{
    \begin{tabular}{lccc}
        \toprule
        \textbf{Model} & \textbf{Input Signals} & \textbf{F1} & \textbf{BLEU} \\
        \midrule
        SimpleTOD  \cite{hosseini2020simple} & Full Input & 35.80 & 20.20 \\
        SimpleTOD  \cite{hosseini2020simple} & Frequent phrases only & 34.33 & 19.63 \\
        \midrule
        GLMP  \cite{wu2019global} & Full Input & 33.79 & 6.22 \\
        GLMP  \cite{wu2019global} & Frequent phrases only & 32.68 & 6.18 \\
        \bottomrule
    \end{tabular}}
    \caption{Performance of two dialogue models under two training  settings: using full input or partial input containing only frequent 
    phrases. F1 and BLEU measure the accuracy of entity prediction and quality of generated response respectively.}
    \label{motivation_table}
\end{table}
\begin{figure}
    \centering
    \includegraphics[width=3.4in]{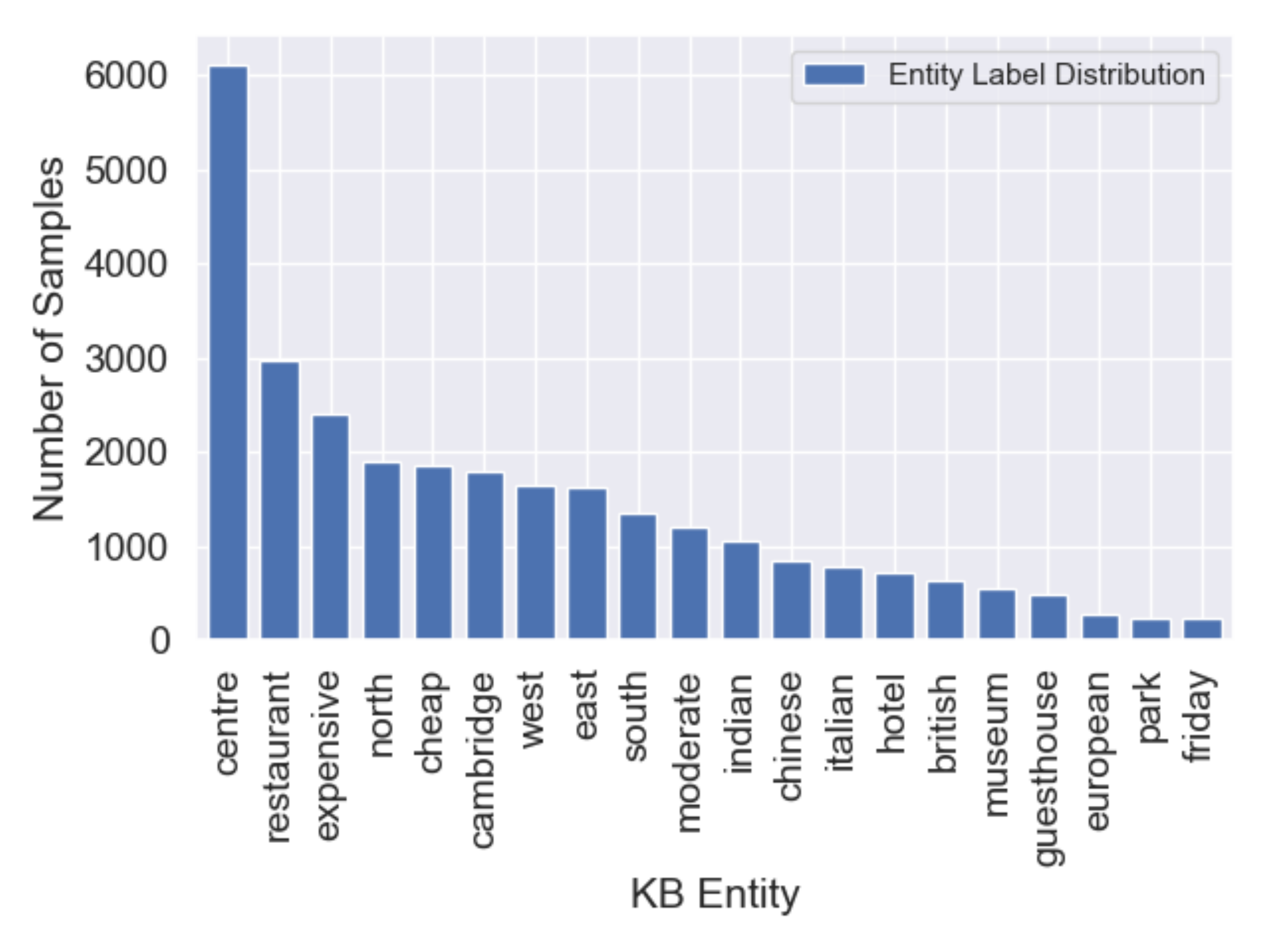}
    \caption{Entity label distribution on benchmark dataset MultiWOZ. X-axis denotes the
    top-20 most frequent entities in the system responses. Y-axis denotes the number of samples
    containing each entity.}
    \label{entity_distribution}
\end{figure}

We experiment with two popular types
of dialogue models based on GPT (SimpleTOD \cite{hosseini2020simple}) and recurrent networks (GLMP \cite{wu2019global}). We evaluate using Entity F1 \cite{eric2017key} and BLEU  \cite{papineni2002bleu} metrics which are the two main metrics 
for assessing the performance of dialogue models. F1 measures the 
accuracy of the system's ability to extract the correct
 entities from the knowledge base, while 
BLEU measures how much word overlap between the system-generated response and the ground truth response; higher score means better performance in both metrics. The
results are shown in Table \ref{motivation_table}. As we can see, both models perform similarly under the two training settings, implying that there are shortcuts in the data (i.e., frequent phrases that correlate strongly with entities and responses), and the model has learned to exploit these cues for the task. Manual analysis reveals that 87\% of the  frequent
phrases do not contain much semantic information: most of them are made up of function words such as \textit{I'm looking for},  \textit{I would like},
\textit{I don't care}, and \textit{That is all}. These results suggest that these models did not solve the task by having any real natural language understanding.

Next we look into class imbalance, another  source of dataset biases \cite{branco-etal-2021-shortcutted}.
We analyze the distribution of KB entities in the system responses, i.e.,  we tally how often each entity appears in the responses in MultiWOZ and plot a histogram of their frequencies in Figure \ref{entity_distribution}.
We find that the entity distribution  is highly skewed, with the top-10 ``head entities'' (i.e., the most frequent entities) 
accounting for approximately 64\% of total occurrences, which means a large portion of the entities are in the (very) long tail of the distribution.
The implication is that a model can simply focus on learning from a small number of head entities to achieve a high performance.
Motivated by this observation, we adapt filtering algorithms to tackle this class imbalance issue, which works by smoothing the distribution so as to encourage our model to learn not only from the head of the distribution but also from the tail.

\begin{figure*}
    \centering
    \includegraphics[width=\textwidth]{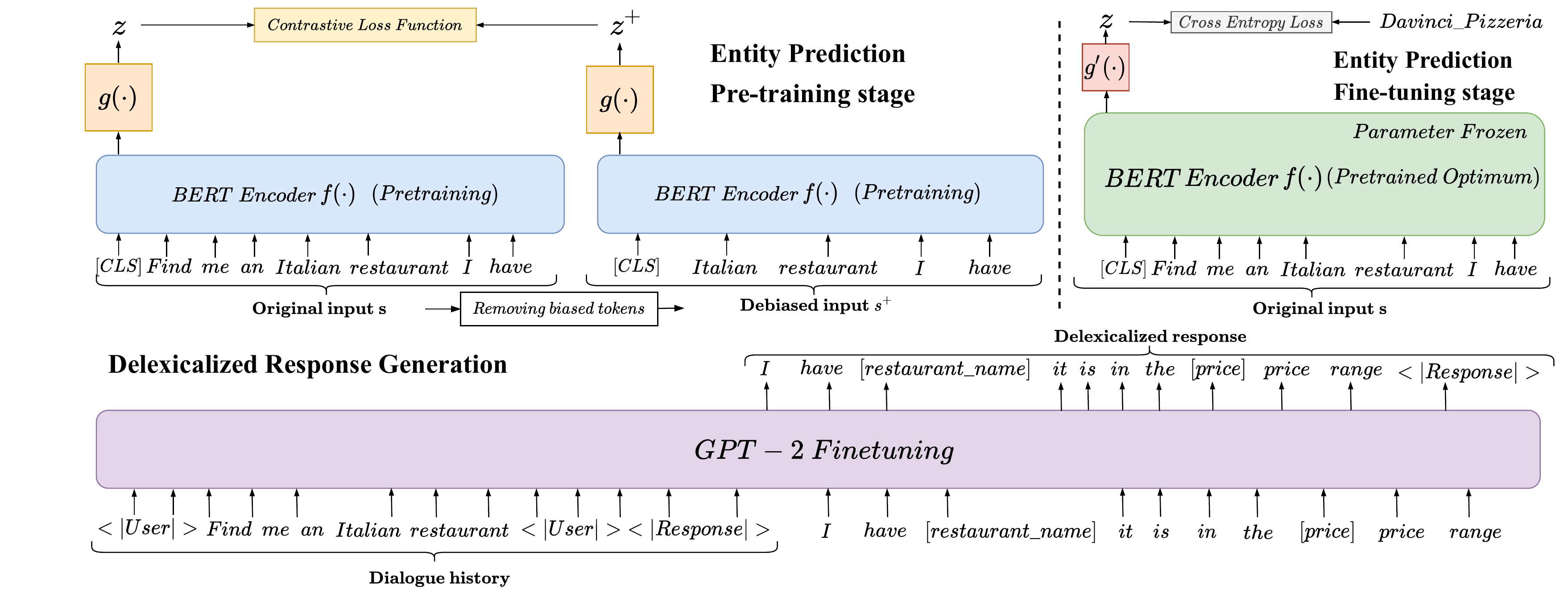}
    \caption{Overview of our proposed approach. The top  shows the 
    two-stage design for entity prediction: BERT pre-training using contrastive loss
    and fine-tuning using cross-entropy loss. The bottom  
    shows the delexicalized response generation by fine-tuning GPT-2.}
    \label{architecture}
\end{figure*}

\section{Model Architecture}
\subsection{Problem Formulation}

We focus on the problem of task-oriented dialogue response generation with external knowledge base (KB). Formally, given the dialogue history $D$ and knowledge base $B$, our goal is to generate the system responses $Y$ in a word-by-word fashion. The probability of the generated responses can be written as:

\begin{equation*}
    \small
    p(Y|D,B) = \prod_{t=1}^{n} p(y_{t}|D,B,y_{1},y_{2},...,y_{t-1})
\end{equation*}

\noindent where $y_{t}$ is the t-th token in the response $Y$.

\subsection{Overview}
We decompose the dialogue generation task into two sub-tasks:  delexicalized response generation and entity prediction.
The delexicalized response is the response where KB entities are substituted
by placeholders to reduce the complexity of the problem through a smaller vocabulary. For example, 
in Figure \ref{architecture}, \textit{Davinci Pizzeria} is 
replaced by ``[restaurant\_name]'' in the response. We follow the delexicalization process proposed in \cite{hosseini2020simple}. 
We employ the two-phase design because it
disentangles the entity prediction task from response generation task, 
allowing us to focus on bias reduction for entity prediction. Our framework uses a
pre-trained autoregressive model (GPT-2) as the response generator and 
a pre-trained bidirectional encoder (BERT) as the entity predictor.
Note that GPT-2 is fine-tuned to generate the delexicalized responses while the
BERT model is fine-tuned to predict entities at every timestep during decoding, and the final response is created by replacing the placeholder tokens (generated by GPT-2) using the predicted entities (by BERT). Figure \ref{architecture} presents the overall architecture. We
first describe how the delexicalized response generation operates in Section \ref{sec:response-generation}
followed by entity prediction  in Section \ref{sec:entity-prediction}. We introduce our debiasing techniques for the entity prediction model in Section \ref{sec:debiased-training}.

\subsection{Delexicalized Response Generation}
\label{sec:response-generation}
We follow \cite{hosseini2020simple}
and fine-tune GPT-2 to 
generate the delexicalized responses. Note that the input is always prefixed with the dialogue history and GPT-2 is fine-tuned via cross-entropy loss to predict the next (single turn) response.%
\footnote{To inform the model about user utterances and system utterances
in the dialogue context, 
we add special tokens $<|user|>$ and $<|system|>$ at the beginning 
and end of user and system utterances for each turn of 
the dialogue history, following \cite{hosseini2020simple}. We also add a symbol $<|response|>$ at the 
end of the dialogue history to indicate the start 
of the response generation.}

\subsection{Entity Prediction}
\label{sec:entity-prediction}
The entity prediction task can be formulated as a multi-class classification problem. The 
goal of the entity prediction module is to predict the correct KB 
entities at each timestep during the response generation process,
given the dialogue context and the generated word tokens before
current timestep. Formally, let D = [$x_1$,$x_2$,...,$x_n$]
be the dialogue history, Y = [$y_1$,$y_2$,...,$y_m$] be the 
ground truth delexicalized response, where
$n$ is the number of tokens in the dialogue history, $m$ the
number of tokens in the response. During training, we fine-tune BERT to 
predict the entity at the $t$-th timestep,
by taking the dialogue history and the generated tokens,
i.e., $\bar{D}$ = [$x_1$,$x_2$,...$x_n$,$y_1$,$y_2$,...,$y_{t-1}$], 
as the input:
\begin{align*}
    {H} &= {BERT}_{{\scriptsize enc}}(\phi^{\scriptsize emb}({\bar{D}})) \\
   z &= g(H_{CLS}) \\
    P &= \text{softmax}(z)
\end{align*}
where $\phi^{\scriptsize emb}$ is the embedding layer of BERT,
$H_{CLS}$ the hidden state of the \texttt{[CLS]} token, $g$ a 
linear layer, and $P$ the probability distribution over the KB entity set.
Note that the KB entity set consists of all KB entities and a special label 
\texttt{[NULL]}, which is used when the token to be predicted at timestep $t$ is not an entity (i.e., normal words).
During inference, we use the delexicalized response generated by GPT-2 as input, and at each time step select the entity with the largest probability produced by BERT as the output.


The delexicalized response generator (GPT-2) and entity predictor (BERT) are trained separately, and at test time we first generate the delexicalized response and then use it as input to the entity predictor to predict the entities at every time step. Once that's done, we lexicalize the response by substituting the placeholder tokens with their corresponding entities to create the final response.

\section{Debiased Training For Robust Entity Prediction}
\label{sec:debiased-training}
Motivated by our preliminary findings a model that uses full input (\textit{I need to find a moderately priced hotel}) performs similarly compared to one that uses filtered input containing only frequent phrases (\textit{I need to}),
we propose to use contrastive learning to encourage the entity predictor (BERT) to focus on the important semantic words, e.g., \textit{find a moderately priced hotel} instead of \textit{I need to}, during representation learning.
We propose three methods leveraging $n$-gram  statistics  ($n = 3$): frequency  (Section \ref{frequent_ngrams}), mutual information
 (Section \ref{mutual_information}), and Jensen-Shannon divergence  (Section \ref{jsd}).

Our approach integrates contrastive learning in the domain-adaptive pre-training stage \cite{Gururangan+:2020}. That is, we take the off-the-shelf pre-trained BERT, and perform another step of pre-training to adapt it to our domain. Conventionally this domain-adaptive pre-training is done using masked language model loss, but we propose to use contrastive loss instead to encourage BERT to learn similar representations/encodings between the full input (\textit{I need to find a moderately priced hotel}) and a \textit{debiased input} (\textit{find a moderately priced hotel}), thereby forcing the model to focus on the semantic words. After this contrastive pre-training, we fine-tune BERT for the entity prediction task as we described in Section \ref{sec:entity-prediction}.

Lastly, we also explore adapting an adversarial filtering algorithm \cite{sakaguchi2019winogrande} to further debias our model; this is detailed in Section \ref{af}.

\subsection{Contrastive Learning}
\label{sec:contrastive-learning}
The core idea of contrastive learning is to learn
representations where positive pairs are embedded in a similar space while negative pairs are pushed apart as much as possible \cite{gao2021simcse,khosla2021supervised,oord2019representation}. 
We follow the
contrastive learning framework in \cite{gao2021simcse} that takes
a set of paired utterances $S$ = $\{(s_i,s_i^{+})\}^{N}_{i=1}$ 
as inputs, where $s_i$ denotes the original input
and $s_i^{+}$ denotes its positive counterparts (i.e., the debiased utterances). It employs 
in-batch negatives and cross-entropy loss for training. Formally, the inputs $s_i$ and 
$s_i^{+}$ are first mapped into feature representations in
vector space as $z_i$ and $z_i^{+}$. In our case we use
 BERT as our encoder to produce the features. The training loss $\mathcal{L}$ for $S$, a minibatch with $N$ pairs of utterances is:
\begin{equation*}
    \mathcal{L} = -\mathop{\mathbb{E}}\limits_{S} \left[ log \frac{e^{\text{sim}(z_i, z_i^{+})/\tau}}{\sum_{j=1}^{N}e^{\text{sim}(z_i, z_j^{+})/\tau}} \right]
    \label{contrastive_loss}
\end{equation*}

\noindent where $\tau$ is a temperature hyperparameter, $sim$ the cosine 
similarity function. 
The critical issue of contrastive learning is
to construct a \textit{meaningful} positive counterpart, which in our case means capturing the semantic bearing words in the original utterance. We next describe three ideas to construct the positive pairs based on $n$-gram statistics.

\subsubsection{Criterion-1: Frequent $n$-grams}\label{frequent_ngrams}
We select the top-10\% $n$-grams according to their frequency in the training data,\footnote{Recall that we perform entity prediction for every timestep in the response, and so for each response we have $m$ training instances, where $m$ is the number of tokens in the response.}  and create positive pairs containing: (1) an original input (dialogue history and response up to timestep $t-1$); and (2) filtered input where frequent $n$-grams are removed. As explained earlier in Section \ref{sec:debiased-training}, this simple approach forces BERT to learn a similar representation between the full input (\textit{I need to find a moderately priced hotel}) and debiased input (\textit{find a moderately priced hotel}) by removing these  $n$-grams directly.

\subsubsection{Criterion-2: Mutual Information}\label{mutual_information}


The previous approach does not consider
the label information (i.e., the entities contained in 
the responses). To incorporate label information, we explore computing mutual information between $n$-grams and the entities. The idea is that we want to discover $n$-grams that  produce strong correlation with entities, which means that BERT is likely to pick them up as shortcuts for prediction. 
Formally:
\begin{equation*}
    I({A};{B}) = \mathop{\sum}\limits_{A,B} p({A},{B}) log \frac{p({A|B)}}{p({A})}
    \label{mutual_info}
\end{equation*}
where ${A}$ is an $n$-gram and ${B}$ a target entity.

We rank all pairs of $n$-grams and entities this way, and select the top-10\% pairs and use their $n$-grams (ignoring the entities) as the candidate set where we remove them in the input to create the positive pairs as before. The detailed algorithm is shown in Algorithm 
\ref{mutual_info_algorithm}.

\subsubsection{Criterion-3: Jensen-Shannon Divergence}\label{jsd}

The previous approach accounts for label (entity) information, but has the limitation where it considers only the \textit{presence} of an $n$-gram with a target entity. Here we extend the approach to also consider the \textit{absence} of the $n$-gram, and what impacts this brings to the appearance of the target entity. To this end we compute the Jensen-Shannon divergence of two probability distributions: (1) entity distribution where an $n$-gram is present in the input ($P$); and (2) entity distribution where an $n$-gram is absent in the input ($Q$). The idea is that an $n$-gram is highly informative (in terms of predicting the entities) if the divergence of the distributions is high, and we want to remove these $n$-grams from the input. Formally:
\begin{equation*}
    JSD({P};{Q}) = \frac{1}{2} \sum_{x \in \mathbb{X}}{P}(x)log\frac{{P}(x)}{{M}(x)} + \frac{1}{2} \sum_{x \in \mathbb{X}}{Q}(x)log\frac{{Q}(x)}{{M}(x)}
    \label{jsd_equation}
\end{equation*}
where ${P}$ and ${Q}$ are two probability distributions over $\mathbb{X}$, ${M}$ = 1/2 (${P}$+${Q}$). Details about the algorithm are shown in Algorithm \ref{jsd_algorithm}. 
As before, we select the top-10\% $n$-grams ranked by the divergence values as the candidate $n$-grams to filter in the input.

\begin{algorithm}
\caption{Mutual information based bias identification}\label{mutual_info_algorithm}
\SetKwInOut{Input}{Input}\SetKwInOut{Output}{Output}
\SetKwFunction{FindNgrams}{FindNgrams}\SetKwFunction{CalcMutualInfo}{CalcMutualInfo}

\Input{dataset $\mathcal{D}$ = ($\mathbf{X}, \mathbf{y}$), $n$-gram size $\textit{n}$, cutoff size $\textit{k}$}
\Output{biased tokens set $\mathcal{S}$}

Initialize $n$-grams distribution $\mathbf{\textit{P}}_{ngrams}$ $\leftarrow$ $\emptyset$\;
Initialize entity distribution $\mathbf{\textit{P}}_{entity}$ $\leftarrow$ $\emptyset$\;
Initialize ngrams-entity joint distribution $\mathbf{\textit{P}}_{joint}$ $\leftarrow$ $\emptyset$\;
Initialize mutual information values $\mathbf{\textit{Q}}$ $\leftarrow$ $\emptyset$\;
\ForEach{$(x, y) \in (\mathbf{X}, \mathbf{y})$}{
    $ngrams$ $\leftarrow$ \FindNgrams{x, n}\;
    \lForEach{element $e$ of the $ngrams$}{Add $e$ to $\mathbf{\textit{P}}_{ngrams}$}
    {Add $y$ to $\mathbf{\textit{P}}_{entity}$}\;
    \lForAll{$e \in ngrams$}{Add ($e, y$) to $\mathbf{\textit{P}}_{joint}$}
}
\ForAll{$(e_1, e_2) \in \mathbf{\textit{P}}_{joint}$}{
    Calculate mutual information $m$ of ($e_1, e_2$) using $\mathbf{\textit{P}}_{ngrams}$, $\mathbf{\textit{P}}_{entity}$, $\mathbf{\textit{P}}_{joint}$ with Equation \ref{mutual_info}\;
    Add ($e_1, e_2, m$) to $Q$\;
}
Select the top-$\textit{k}$ elements $\mathcal{F}$ of $Q$ and Add $\mathcal{F}$($e_1$) to $\mathcal{S}$\;
return $\mathcal{S}$
\end{algorithm}

\begin{algorithm}
\caption{JSD based bias identification}\label{jsd_algorithm}
\SetKwInOut{Input}{Input}\SetKwInOut{Output}{Output}
\SetKwFunction{FindNgrams}{FindNgrams}

\Input{dataset $\mathcal{D}$ = ($\mathbf{X}, \mathbf{y}$), $n$-gram size $\textit{n}$, cutoff size $\textit{k}$}
\Output{biased tokens set $\mathcal{S}$}

Initialize $n$-grams set $\mathbf{\textit{S}}_{ngrams}$ $\leftarrow$ $\emptyset$\;
Initialize positive entity distribution $\mathbf{\textit{P}}_{positive}$ $\leftarrow$ $\emptyset$\;
Initialize negative entity distribution $\mathbf{\textit{P}}_{negative}$ $\leftarrow$ $\emptyset$\;
Initialize JSD values $\mathbf{\textit{Q}}$ $\leftarrow$ $\emptyset$\;
\ForEach{$x \in \mathbf{X}$}{
    $ngrams$ $\leftarrow$ \FindNgrams{x, n}\;
    \lForEach{element $e$ of the $ngrams$}{Add $e$ to $\mathbf{\textit{S}}_{ngrams}$}
}
\ForEach{$(x, y) \in (\mathbf{X}, \mathbf{y})$}{
    \ForEach{$e \in ngrams$}{
        \lIf{$e \in x$}{Add ($e, y$) to $\mathbf{\textit{P}}_{positive}$}
        \lElse{Add ($e, y$) to $\mathbf{\textit{P}}_{negative}$}
    }
}
\ForEach{$e \in \mathbf{\textit{S}}_{ngrams}$}{
    Calculate JSD values $v$ for $e$ using $\mathbf{\textit{P}}_{positive}$, $\mathbf{\textit{P}}_{negative}$ with Equation \ref{jsd_equation}\;
    Add ($e, v$) to $Q$\;
}
Select the top-$\textit{k}$ elements $\mathcal{F}$ of $Q$ and Add $\mathcal{F}$($e$) to $\mathcal{S}$\;
return $\mathcal{S}$
\end{algorithm}


\subsection{Adversarial Filtering}\label{af}
Contrastive pre-training identifies and eliminates biased tokens
at the input representation level. However, this does not change
the entity label distribution, which is another
 source of bias that  makes deep learning models brittle to unseen scenarios \cite{peng2020raddle,liu2021robustness}. To address this, we adapt the adversarial filtering proposed 
by \cite{sakaguchi2019winogrande} to smooth the entity label distribution to prevent the model from learning only from the head of the distribution (frequent entities) but also from the tail of the distribution (rarer entities). The core idea of 
adversarial filtering is to filter out  ``easy'' training examples
--- training instances where their removal doesn't negatively impact the model --- to encourage the model to learn from the  ``hard'' examples, through an 
iterative process utilizing weak linear learners.

During each iteration, we train 100 linear classifiers (logistic regression) on 
a randomly sampled subset (30\%) of training instances.
When the training of each classifier converges, we use it to 
make predictions for the remaining 70\% instances and record their predictions. 
At the end of each iteration, we compute the average prediction accuracy for each instance by calculating the ratio of correct predictions over all classifiers, and filter out instances that have a prediction
accuracy $>= 0.75$ and repeat the process with the remaining instances for the next iteration. The algorithm terminates when less than 500 instances are filtered during one iteration
or when it has reached 100 iterations.
After the filtering process, any instances that are not filtered are used for to \textit{further} fine-tune the entity predictor (BERT). Note that we apply this fine-tuning on the best model (based on validation) from contrastive learning (Section \ref{sec:contrastive-learning}), and following previous studies \cite{tian2020contrastive,chen2020simple,khosla2021supervised} we freeze the BERT parameters and initialise (randomly) a new linear layer.

\begin{table*}[t]
    \setlength{\abovecaptionskip}{-5pt}
    \centering
    \begin{tabular}{lc@{\;\;}cc@{\;\;}cc@{\;\;}cc@{\;\;}cc@{\;\;}c}
        \toprule
         \multirow{2}{*}{$\textbf{Model}$} & \multicolumn{2}{c}{$\textbf{Original}$} & \multicolumn{2}{c}{$\textbf{WP}$} & \multicolumn{2}{c}{$\textbf{WD}$} & \multicolumn{2}{c}{$\textbf{SP}$} & \multicolumn{2}{c}{$\textbf{SI}$} \\
          \cmidrule{2-11}
          & F1 & BLEU & F1 & BLEU & F1 & BLEU & F1 & BLEU & F1 & BLEU \\
          \midrule
          Mem2Seq \cite{mem2seq} & 23.42 & 4.53 & 7.69 & 3.94 & 8.20 & 3.91 & 9.65 & 4.17 & 10.25 & 4.06 \\
          GLMP \cite{wu2019global} & 33.79 & 6.22  & 10.21 & 5.34  & 10.97 & 5.29  & 13.14  & 5.69  & 14.04 & 5.53 \\
          DF-Net \cite{qin-etal-2020-dynamic} &  35.73 & 7.01 & 11.02 & 6.08 & 11.73 & 6.03 &  14.01 & 6.45  & 14.96 & 6.28 \\
          SimpleTOD \cite{hosseini2020simple} & 35.80 & 20.20 & 11.19 & 9.25 & 12.02 & 9.19 & 14.37 & 10.63 & 15.35 & 10.45 \\
          \midrule
          $\textup{Ours (w/o CL, w/ MLM)}^{\diamondsuit}$ & 36.74  & 21.05 & 11.33 & 10.99 & 12.67 & 10.97 & 15.01 & 11.53 & 16.47 & 12.24 \\
          \midrule
          $\textup{Ours (w/ CL, Frequent $n$-grams)}^{\spadesuit}$ & 32.20 & 20.61 & 27.85 & 17.87 & 26.66 & 17.23 & 26.44 & 16.33 & 19.90 & 16.04 \\
          $\textup{Ours (w/ CL, Mutual Information)}^{\spadesuit}$ &  31.86 & 20.04  & 28.84 & 18.38 & 26.98 & 17.36 & 27.14 & 16.60 & 23.60 & 16.89 \\
          $\textup{Ours (w/ CL, Jensen-Shannon Divergence)}^{\spadesuit}$ &  31.50 & 19.42 & 29.15 & 18.96 & 28.13 & 18.33 & 28.95 &  17.81 & 23.96 & 17.37 \\
          \midrule
          $\textup{Ours (w/o CL, w/ AF)}^{\heartsuit}$ & 31.35  & 20.37  & 27.22 & 17.80 & 26.51 & 17.57 & 26.04 & 17.14 & 21.30 &  16.49 \\
          $\textup{Ours (w/ CL, w/ AF)}^{\heartsuit}$ & 30.98  & 19.26  & 29.89 & 19.01 & 29.28 & 18.93 & 29.20 & 18.06 & 28.69 & 17.59 \\
          \bottomrule
    \end{tabular}
    \caption{Adversarial attack results. All the models are trained on the original (unperturbed) MultiWOZ data using all domains. ``Original'' denote the original MultiWOZ test set, while ``WP'', ``WD'', ``SP'' and ``SI'' denote adversarial test sets created through word paraphrasing, word deletion, sentence paraphrasing and sentence insetion respectively, using NlpAug \cite{ma2019nlpaug}. $\diamondsuit$: Our vanilla model without contrastive learning or adversarial filtering; $\spadesuit$: our model with contrastive learning; $\heartsuit$: our model with adversarial filtering.}
    \label{adversarial_attack_results}
\end{table*}

\section{Experiments}
To verify the effectiveness of our proposed debiasing approach, 
we conduct a comprehensive study comparing our model
against a number of benchmark systems. Our experiments include 
cross-domain/dataset generalization test, adversarial 
 samples (created by distorting words and sentences), and utterances featuring unseen $n$-grams.


\subsection{Datasets and Metrics}
We use MultiWOZ \cite{eric2019multiwoz} as the main dialogue 
dataset for our experiments. Specifically,
we use version 2.2 of the dataset \cite{zang2020multiwoz}  which fixes a number of annotation 
errors and disallows slots with a large number of values to improve data quality. For evaluation metrics, we use the same BLEU and Entity F1 measures that we used in our preliminary investgation (Section \ref{sec:preliminary}).

\subsection{Baselines}
We compare our model against the following state-of-the-art benchmark systems:

\begin{itemize}
    \item {{Mem2Seq}} \cite{mem2seq}:  employs a recurrent network-based decoder to generate
        system responses and utilize memory networks to store the 
        KB and copy KB entities from memory via pointer mechanism. The
        decoder are jointly trained with memory networks end-to-end
        by maximizing the likelihood of the final system responses.
    \item {{GLMP}} \cite{wu2019global}: 
        employs a global-to-local pointer mechanism over the standard 
        memory networks architecture for improving KB retrieval accuracy during
        response generation. The global pointer is supervised by additional
        training signals extracted from the standard system responses.
    \item {{DF-Net}} \cite{qin-etal-2020-dynamic}: utilizes 
        a shared-private architecture to capture both domain-specific and 
        domain-general knowledge to improve the model transferability.
    \item {{SimpleTOD}} \cite{hosseini2020simple}: a causal language 
        model based on GPT-2 trained on several task-oriented dialogue 
        sub-tasks including dialogue state tracking, action prediction and 
        response generation. It exploits additional training signals 
        such as dialogue states and system acts compared to other systems.
\end{itemize}

\subsection{Implementation Details}
For delexicalized response generation, we use pretrained \texttt{gpt2}.\footnote{\url{https://huggingface.co/gpt2}.}
We use the default hyper-parameter configuration,
except for learning rate and batch size where we optimise via grid search. The learning rate is selected from 
\{\textit{$1e^{-3}$},\textit{$1e^{-4}$},\textit{$1e^{-5}$},
\textit{$2e^{-4}$},\textit{$2e^{-4}$},\textit{$2e^{-4}$}\} and 
batch size from \{2,4,8,16,32\} based on the best validation 
performance.

For entity prediction, we 
use pretrained \texttt{bert-base-uncased}.\footnote{\url{https://huggingface.co/bert-base-uncased}.} During contrastive pre-training stage, the learning 
rate is selected from \{\textit{$1e^{-3}$},\textit{$1e^{-4}$},
\textit{$1e^{-5}$},\textit{$1e^{-6}$}\} 
and batch size  from \{2,4,8,16,32\} 
using grid search based on validation performance. We pre-train
BERT with contrastive learning for 20 epochs.
The model with the best validation performance (minimum loss) is 
used for fine-tuning for entity prediction. During fine-tuning
stage, the learning rate is selected from 
\{\textit{$1e^{-3}$},\textit{$1e^{-4}$},\textit{$1e^{-5}$},\textit{$2e^{-4}$},
\textit{$2e^{-4}$},\textit{$2e^{-4}$}\} and the batch size  
from \{2,4,8,16,32\}. We use Adam \cite{kingma2014adam} as the
optimizer.
In terms of $n$-gram order, $n =3$ (trigram).

We run all experiments five times using 
different random seeds and report the average and standard deviation. 
All the models are trained on a single GeForce RTX 2080 Ti GPU and the 
training of both components (response generator and entity predictor) takes approximately one day.

\begin{table*}[t]
    \setlength{\abovecaptionskip}{-3pt}
    \centering
    \begin{adjustbox}{max width=1.0\textwidth}
    \begin{tabular}{lcccccc}
        \toprule
        \multirow{2}{*}{$\textbf{Model}$} & $\textbf{All}$ & $\textbf{All}$ & $\textbf{Restaurant}$ & $\textbf{Hotel}$ & $\textbf{Attraction}$ & $\textbf{Training}$ \\
         & $\textbf{F1}$ & $\textbf{BLEU}$ & $\textbf{F1}$ & $\textbf{F1}$ & $\textbf{F1}$ & $\textbf{F1}$ \\
        \midrule
        Mem2Seq \cite{mem2seq} & 10.38 ($\pm$0.28) & 4.27 ($\pm$0.79) & 13.51 ($\pm$0.13) & 5.59 ($\pm$0.26) & 14.85 ($\pm$0.30) & 8.66 ($\pm$0.50) \\
        GLMP \cite{wu2019global} & 14.25 ($\pm$0.20) & 5.84 ($\pm$0.05) & 18.93 ($\pm$0.42) & 7.06 ($\pm$0.47) & 20.95 ($\pm$0.29) & 11.67 ($\pm$0.20) \\
        DF-Net \cite{qin-etal-2020-dynamic} & 15.17 ($\pm$0.31) & 6.61 ($\pm$0.41) & 20.10 ($\pm$0.20) & 7.61 ($\pm$0.30) & 22.23 ($\pm$0.43) & 12.46 ($\pm$0.29) \\
        SimpleTOD \cite{hosseini2020simple} & 15.57 ($\pm$0.17) & 12.79 ($\pm$0.25) & 20.64 ($\pm$0.12) & 7.78 ($\pm$0.10) & 22.83 ($\pm$0.30) & 12.78 ($\pm$0.14) \\
        \midrule
        $\textup{Ours (w/o CL, w/ MLM)}^{\diamondsuit}$ & 15.82 ($\pm$0.46) & 14.36 ($\pm$0.21) & 18.04 ($\pm$0.09) & 8.68 ($\pm$0.11) & 23.16 ($\pm$0.19) & 14.49 ($\pm$0.47) \\
        \midrule
        $\textup{Ours (w/ CL, Frequent $n$-grams)}^{\spadesuit}$ & 20.33 ($\pm$0.15) & 17.17 ($\pm$0.15) & 22.29 ($\pm$0.09) & 10.23 ($\pm$0.21) & 20.82 ($\pm$0.12) & 9.65 ($\pm$0.02) \\
        $\textup{Ours (w/ CL, Mutual Information)}^{\spadesuit}$ & 21.25 ($\pm$0.04) & 17.86 ($\pm$0.05) & 20.80 ($\pm$0.05) & 12.91 ($\pm$0.31) & 26.05 ($\pm$0.28) & 13.72 ($\pm$0.16) \\
        $\textup{Ours (w/ CL, Jensen-Shannon Divergence)}^{\spadesuit}$ & 26.42 ($\pm$0.14) & 21.83 ($\pm$0.07) & $\textbf{28.46 ($\pm$0.15)}$ & 14.72 ($\pm$0.07) & 27.07 ($\pm$0.12) & 15.64 ($\pm$0.04) \\
        \midrule
        $\textup{Ours (w/o CL, w/ AF)}^{\heartsuit}$ & 20.26 ($\pm$0.11) & 15.53 ($\pm$0.25) & 20.30 ($\pm$0.08) & 10.04 ($\pm$0.06) & 24.31 ($\pm$0.04) & 14.22 ($\pm$0.19) \\
        $\textup{Ours (w/ CL, w/ AF)}^{\heartsuit}$ & $\textbf{29.80 ($\pm$0.27)}$ & $\textbf{22.05 ($\pm$0.15)}$ & 27.51 ($\pm$0.37) & $\textbf{22.59 ($\pm$0.17)}$ & $\textbf{40.07 ($\pm$0.09)}$ & $\textbf{15.72 ($\pm$0.28)}$\\
        \bottomrule
    \end{tabular}
    \end{adjustbox}
    \caption{Unseen utterances generalization test results. All the training and inference are done on the new  train/test splits (see Section \ref{unseen_utterance}) created to reduce $n$-gram overlap between training and test data. ``F1 (All)'': testing F1 using all the domains; ``BLEU (All)'': testing BLEU using all the domains; ``Training F1'': F1 results on training partition using all the domains. We run each experiment five times with different random seeds and report the average results with standard deviation (in parenthesis).}
    \label{unseen_utterance_results}
\end{table*}

\subsection{Adversarial Attack Results}
Language variety is one of the key features of human 
languages \cite{ganhotra2020effects}, i.e., we tend to express 
the same meaning using different words. In real-world 
situations, users may use very different expressions 
than those in the training data. To test model
robustness under such situations, we perform several 
perturbations on user utterances in the original test set to construct adversarial 
test examples. We use the widely-used 
NlpAug  library \cite{ma2019nlpaug} to augment 
the ``regular'' user utterances to generate four adversarial test sets through: 
word paraphrasing (WP), word deletion (WD), 
sentence paraphrasing (SP), and sentence insertion (SI).
All the hyper-parameters of the augmentation tool NlpAug are 
kept to their default. We train all systems (benchmark and ours) using the \textit{original} MultiWOZ 
and test them on both the original test 
set and adversarial test sets. Results 
are shown in Table \ref{adversarial_attack_results} \footnote{Due to space limit, we report average 
performance for adversarial attack experiment. For other experiments, we report both average and standard deviation.}.

Our model has several variants: (1) vanilla without any debiasing, noting that it still has domain adaptive pre-training using the masked language model loss ($\diamondsuit$); (2) with contrastive loss for domain-adaptive pre-training ($\spadesuit$); and (3) with adversarial filtering, applied with or without contrastive pre-training ($\heartsuit$). The reason why our vanilla model has masked language model pre-training is that we need to understand that when we introduce contrastive pre-training, any performance gain is attributed to the contrastive learning objective rather than the domain adaptive pre-training step.

Looking at the original test set (``Original''), among the benchmark systems SimpleTOD is the best model, and our vanilla model ($\diamondsuit$) performs similarly (marginally better F1 but lower BLEU). Introducing contrastive learning ($\spadesuit$) and adversarial filtering ($\heartsuit$) somewhat degrades the entity prediction performance (F1), although the quality of the generated response (BLEU) is less impacted. Moving on to the adversarial test sets, all benchmark systems and our vanilla model observe severe performance degradation: F1 drops by over 20 points and BLEU by 10 points for most systems, suggesting that these models are not robust against perturbed inputs. Our systems with contrastive learning and/or adversarial filtering, on the other hand, look promising: the drop is substantially less severe, 2-3 points in terms of F1 and BLEU. Interestingly, we also see that SI appears to be the most challenging test set as its performance is lowest. Comparing the three different criteria for ranking $n$-grams (frequent $n$-gram, mutual information and Jensen-Shannon divergence), Jensen-Shannon divergence appears to have the upper hand, suggesting that label information and both the presence \textit{and} absence of an $n$-gram is important for uncovering shortcuts in the data.

\begin{table*}[]
    \centering
    \begin{adjustbox}{max width=1.0\textwidth}
    \begin{tabular}{lc@{\;\;}cc@{\;\;}cc@{\;\;}cc@{\;\;}c}
        \toprule
        \multirow{3}{*}{\textbf{Model}} & \multicolumn{2}{c}{$\textbf{\textit{Hotel,Attraction,Train}}$} & \multicolumn{2}{c}{$\textbf{\textit{Restaurant,Attraction,Train}}$} & \multicolumn{2}{c}{$\textbf{\textit{Restaurant,Hotel,Train}}$} & \multicolumn{2}{c}{$\textbf{\textit{Restaurant,Hotel,Attraction}}$} \\
        & \multicolumn{2}{c}{$\rightarrow$ $\textbf{\textit{Restaurant}}$} & \multicolumn{2}{c}{$\rightarrow$ $\textbf{\textit{Hotel}}$} & \multicolumn{2}{c}{$\rightarrow$ $\textbf{\textit{Attraction}}$} & \multicolumn{2}{c}{$\rightarrow$ $\textbf{\textit{Train}}$} \\
        \cmidrule{2-9}
         & F1 & BLEU & F1 & BLEU & F1 & BLEU & F1 & BLEU \\
         \cmidrule{1-9}
         Mem2Seq \cite{mem2seq} & 9.70 ($\pm$0.04) & 4.88 ($\pm$0.56) & 2.87 ($\pm$0.30) & 3.89 ($\pm$0.48) & 6.97 ($\pm$0.16) & 3.67 ($\pm$0.66) & 2.89 ($\pm$0.08) & 3.03 ($\pm$0.05) \\
         GLMP \cite{wu2019global} & 13.22 ($\pm$0.36) & 6.75 ($\pm$0.09) & 2.98 ($\pm$0.47) & 5.26 ($\pm$0.51) & 9.13 ($\pm$0.40) & 4.94 ($\pm$0.11) & 3.00 ($\pm$0.19) & 3.98 ($\pm$0.26) \\
         DF-Net \cite{qin-etal-2020-dynamic} & 14.09 ($\pm$0.69) & 7.57 ($\pm$0.04) & 3.32 ($\pm$0.38) & 5.99 ($\pm$0.49) & 9.79 ($\pm$0.58) & 5.66 ($\pm$0.10) & 3.34 ($\pm$0.24) & 4.65 ($\pm$0.52) \\
         SimpleTOD \cite{hosseini2020simple} & 14.46 ($\pm$0.47) & 13.78 ($\pm$0.40) & 3.36 ($\pm$0.08) & 11.16 ($\pm$0.27) & 10.03 ($\pm$0.15) & 11.82 ($\pm$0.09) & 3.38 ($\pm$0.45) & 13.77 ($\pm$0.39) \\
         \cmidrule{1-9}
         $\textup{Ours (w/o CL, w/ MLM)}^{\diamondsuit}$ & 14.95 ($\pm$0.35) & 14.23 ($\pm$0.07) & 5.80 ($\pm$0.02) & 13.81 ($\pm$0.11) & 10.31 ($\pm$0.19) & 12.93 ($\pm$0.42) & 3.77 ($\pm$0.34) & 14.91 ($\pm$0.03) \\
         \cmidrule{1-9}
         $\textup{Ours (w/ CL, Frequent $n$-grams)}^{\spadesuit}$ & 19.09 ($\pm$0.20) & 15.27 ($\pm$0.07) & 12.34 ($\pm$0.27) & 14.56 ($\pm$0.09) & 13.62 ($\pm$0.14) & 14.32 ($\pm$0.26) & 10.25 ($\pm$0.41) & 14.93 ($\pm$0.06) \\
         $\textup{Ours (w/ CL, Mutual Information)}^{\spadesuit}$ & 22.24 ($\pm$0.19) & 15.33 ($\pm$0.26) & 14.85 ($\pm$0.35) & 14.60 ($\pm$0.14) & 15.05 ($\pm$0.26) & 15.93 ($\pm$0.44) & 11.76 ($\pm$0.08) & 17.46 ($\pm$0.47) \\
         $\textup{Ours (w/ CL, Jensen-Shannon Divergence)}^{\spadesuit}$ & 23.14 ($\pm$0.25) & 15.68 ($\pm$0.13) & 18.01 ($\pm$0.38) & 15.01 ($\pm$0.25) & 19.61 ($\pm$0.47) & 16.38 ($\pm$0.07) & 12.48 ($\pm$0.32) & 18.86 ($\pm$0.44) \\
         \cmidrule{1-9}
         $\textup{Ours (w/o CL, w/ AF)}^{\heartsuit}$ & 21.30 ($\pm$0.17) & 16.35 ($\pm$0.14) & 15.26 ($\pm$0.33) & 14.96 ($\pm$0.48) & 16.76 ($\pm$0.41) & 16.12 ($\pm$0.18) & 10.86 ($\pm$0.31) & 16.30 ($\pm$0.11) \\
         $\textup{Ours (w/ CL, w/ AF)}^{\heartsuit}$ & $\textbf{25.13 ($\pm$0.29)}$ & $\textbf{17.93 ($\pm$0.35)}$ & $\textbf{23.86 ($\pm$0.17)}$ & $\textbf{15.83 ($\pm$0.12)}$ & $\textbf{21.37 ($\pm$0.20)}$ & $\textbf{16.90 ($\pm$0.09)}$ & $\textbf{14.43 ($\pm$0.13)}$ & $\textbf{19.03 ($\pm$0.08)}$ \\
         \bottomrule
    \end{tabular}
    \end{adjustbox}
    \caption{Cross-domain generalization results. ``X$\rightarrow$Y'': X denote the training domain(s) and Y the test domain.}
    \label{cross_domain_results}
\end{table*}

\begin{table*}[]
 \centering
 \begin{tabular}{lcccc}
     \toprule
      \multirow{2}{*}{$\textbf{Model}$} & \multicolumn{2}{c}{$\textit{\textbf{MultiWOZ$\to$SMD}}$}  &  \multicolumn{2}{c}{$\textit{\textbf{MultiWOZ$\to$SGD}}$}  \\
      \cmidrule{2-5}
      & F1 & BLEU & F1 & BLEU \\
      \midrule
      Mem2Seq \cite{mem2seq} & 9.06 ($\pm$0.49) & 8.01 ($\pm$0.33) & 4.34 ($\pm$0.28) & 3.48 ($\pm$0.63) \\
      GLMP \cite{wu2019global} & 12.25 ($\pm$0.21) & 11.45 ($\pm$0.74) & 5.18 ($\pm$0.47) & 4.66 ($\pm$0.38) \\
      DF-Net \cite{qin-etal-2020-dynamic} & 13.07 ($\pm$0.71) & 12.51 ($\pm$0.57) & 5.63 ($\pm$0.79) & 5.36 ($\pm$0.94) \\
      SimpleTOD \cite{hosseini2020simple} & 13.41 ($\pm$0.41) & 13.87 ($\pm$0.37) & 5.75 ($\pm$0.81) & 7.51 ($\pm$0.29) \\
      \midrule
      $\textup{Ours (w/o CL, w/ MLM)}^{\clubsuit}$ & 13.57 ($\pm$0.44) & 14.45 ($\pm$0.69) & 6.06 ($\pm$0.77) & 8.94 ($\pm$0.76) \\
      \midrule
      $\textup{Ours (w/ CL, Frequent $n$-grams)}^{\spadesuit}$ & 19.01 ($\pm$0.70) & 15.25 ($\pm$0.44) & 9.44 ($\pm$0.78) & 10.07 ($\pm$0.73) \\
      $\textup{Ours (w/ CL, Mutual Information)}^{\spadesuit}$ & 20.32 ($\pm$0.42) & 15.44 ($\pm$0.15) & 9.79 ($\pm$0.81) & 10.38 ($\pm$0.76) \\
      $\textup{Ours (w/ CL, Jensen-Shannon Divergence)}^{\spadesuit}$ & 21.80 ($\pm$0.68) & 15.70 ($\pm$0.67) & 11.23 ($\pm$0.14) & 10.68 ($\pm$0.07) \\
      \midrule
      $\textup{Ours (w/o CL, w/ AF)}^{\heartsuit}$ & 20.44 ($\pm$0.39) & 15.43 ($\pm$0.62) & 9.73 ($\pm$0.04) & 10.25 ($\pm$0.11) \\
      $\textup{Ours (w/ CL, w/ AF)}^{\diamondsuit}$ & 23.79 ($\pm$0.49) & 18.25 ($\pm$0.23) & 11.41 ($\pm$0.21) & 10.99 ($\pm$0.05) \\
      \bottomrule
 \end{tabular}
 \caption{Cross-dataset generalization results. ``A$\rightarrow$B'': a model is trained using train partition of dataset A and evaluated on the test partition of dataset B in a zero-shot manner.}
 \label{cross_dataset_results}
\end{table*}

\subsection{Unseen Utterances Generalization Results}\label{unseen_utterance}

To test our model's generalization capability under unseen 
scenario, we construct a new MultiWOZ split that aims to reduce 
 $n$-gram overlap between training 
and test data. We first collect all $n$-gram types in the full data (training $+$ test) and remove low frequency ($<10$) $n$-grams , and then create two sets of $n$-grams based on their frequencies: ``train'' which contains the most frequent (70\%) $n$-gram types and ``test'' for the remaining (30\%). These two sets will decide whether an instance will be assigned to the training or test partition. That is, we
iterate each instance from the full data and put it to the training partition if it only contains ``train'' $n$-gram, or the test partition if it has only ``test'' $n$-gram. \footnote{For instances containing both ``train'' $n$-grams and ``test'' $n$-grams, we put them to the training partition or test partition according to
the majority type. If the numbers of ``train'' $n$-grams and ``test'' $n$-grams are equal for an instance, we put it to the training partition.}

Empirically, in the original MultiWOZ split the $n$-gram overlap ratio is  82.75\%; our new split reduces this to 51.2\%. This means 
 that during testing, a model using our split will be exposed to 
 utterances with more unseen phrases, and if the model exploits 
the spurious cues ($n$-grams) in the input it will likely to perform poorly under 
this new split, as these cues are more likely to be absent.

We train all models using the new training partition
 and test them on the new test partition and present the results in Table \ref{unseen_utterance_results}.
Looking at the models without debiasing, we find a similar observation where SimpleTOD and our vanilla model ($\diamondsuit$) are the best performing models over different domains. When we introduce contrastive learning ($\spadesuit$) and adversarial filtering ($\heartsuit$), we see an improvement over all domains, with the best variant that combines both (``w/ CL, w/ AF'') improving over the vanilla model  by a large margin, about 14\% in Entity F1 and 8\% in BLEU. As before, using Jensen-Shannon divergence as the criterion for ranking $n$-grams turns out to be the best approach. Contrastive learning and adversarial filtering seem to provide complementary signal based on this experiment, as combining them both produces substantially better performance.

To understand how these two methods complement each other, we present F1 performance in the training partition (last column in Table \ref{unseen_utterance_results}): here we see that the models that are not debiased have similar training and test performance, while the contrastive models ($\spadesuit$) have a much lower training performance, suggesting there is underfitting. Adversarial filtering, however, does not suffer from this problem (``w/o CL, w/ AF''), as it only removes training instances that do not negatively impact training accuracy. Interestingly, when we incorporate adversarial filtering to the contrastive model (``w/ CL, w/ AF''),  this underfitting problem is corrected, showing their complementarity.

\subsection{Cross-domain Generalization Results}

We now test cross-domain generalization, where a model is tested using a domain that is not in the training data.
 
We use the ``leave-one-out'' 
strategy for this, where a model is trained using all except one domain and  tested using that unseen domain. We present the results in Table \ref{cross_domain_results}. 

We see similar observations here. Without any debiasing, SimpleTOD and our vanilla have the best performances. When we incorporate contrastive learning and adversarial filtering, we see a strong improvement in terms of model robustness. As before, the best variant is one that combines both, and when compared to the vanilla model it improves F1 by 10--18 and BLEU by 2--5 points depending on the test domain. For contrastive learning, Jensen-Shanon divergence is again the best criterion for selecting $n$-grams. Adversarial filtering by itself is also fairly effective, although not as effective as the best contrastive model.

 \subsection{Cross-dataset Results}

We now test the hardest setting: cross-dataset generalization. If a model ``overfits'' a dataset and relies on spurious correlations to perform a task, it will likely to perform very poorly in a new dataset of the same task. In this experiment, we train systems on MultiWOZ, and 
 test them on two other popular 
 datasets for task-oriented dialogues: SMD \cite{eric2017key}
 and SGD \cite{rastogi2020scalable}.  For SGD, it doesn't 
 have a database like  SMD and MultiWOZ. Following \cite{rastogi2020scalable}, 
 we collect the returned entities from the API queries during each 
 dialogue as the database records to mimic the data settings of SMD and 
MultiWOZ. 

The results are shown in Table \ref{cross_dataset_results}. Both contrastive learning and adversarial filtering are effective methods to improve model robustness. The best contrastive model (``w/ Jensen-Shanon divergence'') improves F1 by 5--8 and BLEU by 1--2 points when compared to the vanilla model. Adversarial filtering by itself is also effective, although the improvement is marginally smaller compared to the best contrastive model. Once again, combining both produces the best performance. SGD is noticeably a harder dataset here, where F1 performance of all models is about half of SMD's.

All in all these generalization tests reveal strikingly similar observations. To summarize: (1) our vanilla model that decomposes task-oriented dialogue generation into delexicalized response generation and entity prediction performs competitively with benchmark systems; (2) for contrastive learning, Jensen-Shannon divergence is consistently the best performer for ranking $n$-grams, implying that it is important to consider both the absence and presence of $n$-grams when determining their correlations with the labels; and (3) contrastive learning and adversarial filtering complement each other, and the most robust model is produced by incorporating both methods. Note that although our experiments focus on task-oriented dialogue, our methodology can be adapted to other NLP tasks without much difficulty. We believe that our contrastive pre-training method is likely useful to reduce general NLP data artifacts, as spurious correlations between words and labels are likely to surface during data development.

\begin{figure}
    \setlength{\abovecaptionskip}{-2pt}
    \centering
    \includegraphics[width=2.7in]{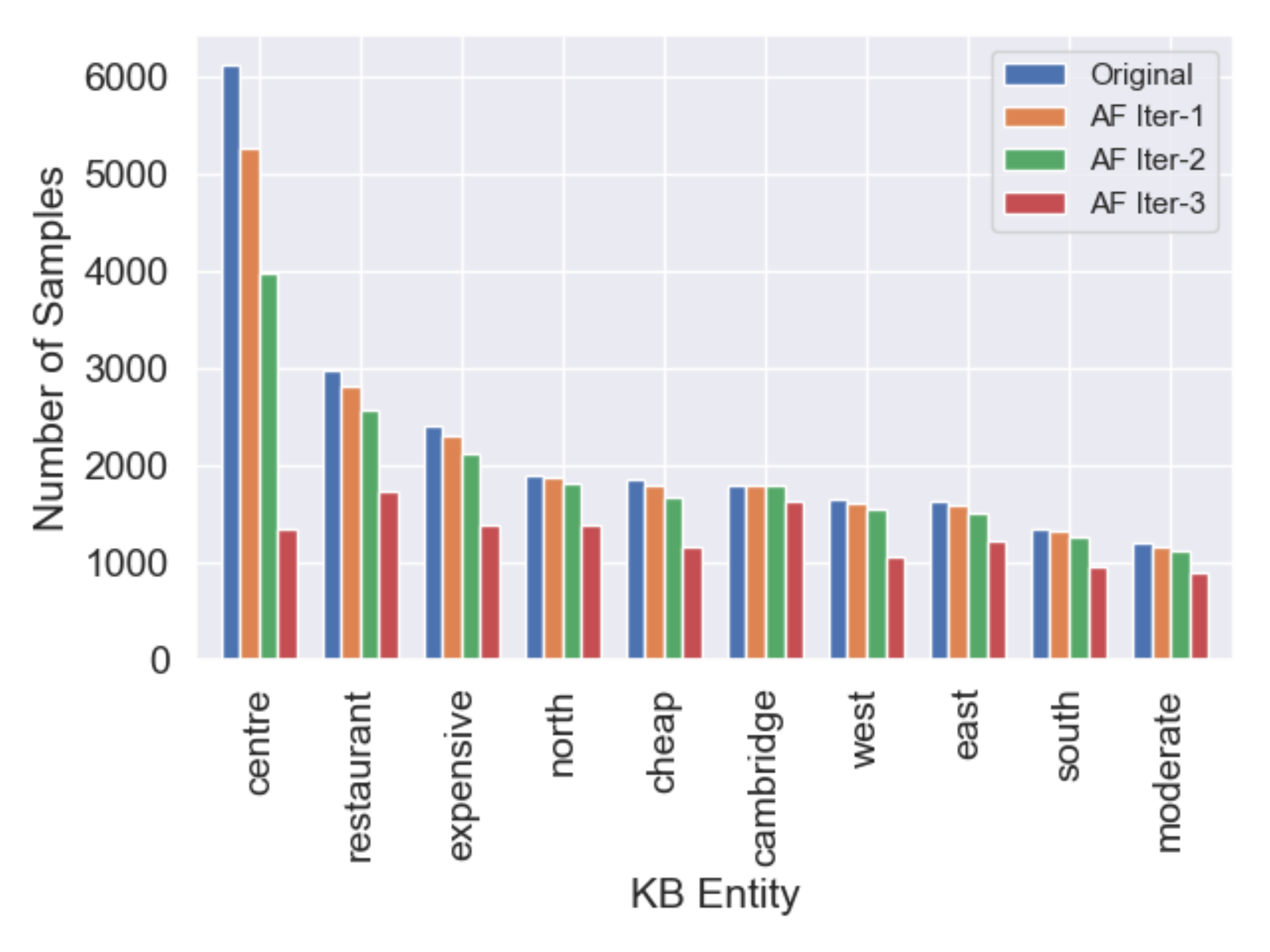}
    \caption{Change in entity frequency over different iterations of adversarial filtering. X-axis shows
    the 10 most frequent entity labels in the response. Y-axis shows the frequency of training instances for an entity. ``Original'': Entity frequency without any filtering; ``AF Iter-k'': Entity frequency after k iterations of filtering.}
    \label{af_dynamics}
\end{figure}

\subsection{Adversarial Filtering Dynamics Analysis}

The adversarial filtering
iteratively detects the easy training instances  and 
remove them. A natural question to ask is: 
why does this filtering process make the model 
more robust? To answer this question, we perform an analysis on 
the dynamics of the iterative process where we analyzed the number 
of instances containing a particular entity. We select the top-10 most frequent entities in the response and monitor their change in frequency over the iterations and present the results in 
Figure \ref{af_dynamics}. As we can see from the figure, the distribution 
over these 10 entities has become  
``flatter'' after the third iteration (red bars). The more frequent an entity is, the more instances are removed: at the extreme, the most frequent entity (\textit{centre}) has lost almost 5000 instances after 3 iterations of filtering.
Intuitively, we believe the more balanced distribution disincentivizes the model to focus on shortcuts that produce the frequent labels (entities), resulting in a model that learns generalisable patterns from a larger set of entities in the tail of the distribution.


\section{Conclusion}
In this work, we investigate data artifacts in MultiWOZ, a popular task-oriented dialogue 
dataset. We find that a model that uses full input for training performs similarly to a variant that uses partial input containing only frequent phrases,  suggesting that there are data artifacts and the model uses them as shortcuts for the task.
Motivated by this analysis, we propose a contrastive learning objective to debias task-oriented dialogue models by encouraging 
models to ignore these frequent phrases to focus on semantic words in the input. We 
also adapt adversarial filtering to our task to further improve model robustness. We conduct a 
series of generalization experiments, testing our method and a number of
state-of-the-art benchmarks. Experimental results show that contrastive learning and adversarial filtering complement each other, and combining both produces the most robust dialogue model.

\bibliographystyle{ACM-Reference-Format}
\bibliography{sample-base,custom}

\appendix

\end{document}